\documentclass[journal]{IEEEtran}
\usepackage{times}
\usepackage{epsfig}
\usepackage{graphicx}
\usepackage{amsmath}
\usepackage{amssymb}
\usepackage{caption}
\usepackage{subcaption}
\usepackage{comment}
\usepackage{adjustbox}
\usepackage{colortbl}
\usepackage{multirow}
\usepackage[dvipsnames]{xcolor}
\usepackage{booktabs}
\usepackage{stmaryrd}
\usepackage{multirow}
\usepackage{enumerate}

\usepackage[pagebackref=true,breaklinks=true,letterpaper=true,colorlinks,bookmarks=false]{hyperref}

\begin{document}

\title{Anti-aliasing Deep Image Classifiers using Novel Depth Adaptive Blurring and Activation Function}

\author{Md~Tahmid~Hossain,~\IEEEmembership{}
        Shyh~Wei~Teng,~\IEEEmembership{}
        Ferdous~Sohel,~\IEEEmembership{Senior~Member,~IEEE,}
        Guojun~Lu,~\IEEEmembership{Senior~Member,~IEEE}
\thanks{Md Tahmid Hossain, Shyh Wei Teng, and Guojun Lu are with the School of Engineering, Information Technology and Physical Sciences, Federation University Australia, Gippsland Campus Churchill, VIC 3842, Australia. E-mail: (mt.hossain,shyh.wei.teng,guojun.lu)@federation.edu.au}
\thanks{Ferdous Sohel is with the Discipline of Information Technology , Murdoch University, Perth, WA 6150, Australia. E-mail: f.sohel@murdoch.edu.au}}


\maketitle

\begin{abstract}
     Deep convolutional networks are vulnerable to image translation or shift, partly due to common down-sampling layers, e.g., max-pooling and strided convolution. These operations violate the Nyquist sampling rate and cause aliasing. The textbook solution is low-pass filtering (blurring) before down-sampling, which can benefit deep networks as well. Even so, non-linearity units, such as ReLU, often re-introduce the problem, suggesting that blurring alone may not suffice. In this work, first, we analyse deep features with Fourier transform and show that Depth Adaptive Blurring is more effective, as opposed to monotonic blurring. To this end, we outline how this can replace existing down-sampling methods. Second, we introduce a novel activation function -- with a built-in low pass filter, to keep the problem from reappearing. From experiments, we observe generalisation on other forms of transformations and corruptions as well, e.g., rotation, scale, and noise. We evaluate our method under three challenging settings: (1) a variety of image translations; (2) adversarial attacks -- both $\ell_{p}$ bounded and unbounded; and (3) data corruptions and perturbations. In each setting, our method achieves state-of-the-art results and improves clean accuracy on various benchmark datasets.
\end{abstract}


\IEEEpeerreviewmaketitle

\section{Introduction}

\IEEEPARstart{C}{onvolutional} Neural Networks (CNNs) have achieved state-of-the-art results in a wide range of vision tasks, including large-scale image classification \cite{imagenet1}. To understand the extent of CNN's robustness, researchers have started looking beyond performance on i.i.d. (independent and identically distributed) datasets only. Recent studies reveal that CNNs are vulnerable to subtle changes in input, e.g., simple one pixel shift \cite{Azulay1,zhang2019making,mairal1}, quasi-imperceptible noise \cite{akaram1,icccn1,HendrycksALP1,hossain1,zhou1}, blur \cite{qomex1,augmix}, and adversarial attacks \cite{Mohapatra1,Engstrom1,Engstrom2}. This is concerning for real-world multi-media applications, where the i.i.d. assumption does not always hold. For instance, an autonomous car should not flip predictions for the same object between consecutive video frames due to marginal spatial shift or image noise \cite{sundaramoorthi2019,manfredi2020}. Recently, aliasing has been identified as one of the main reasons behind CNN's lack of robustness -- especially against small image transformations, such as shift \cite{Azulay1, zhang2019making}. 

\begin{figure}
\begin{center}
   \includegraphics[width=.85\linewidth]{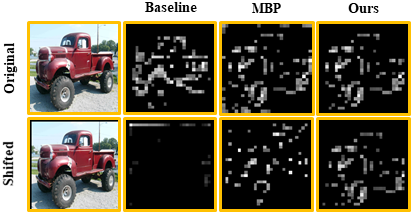}
\end{center}
   \caption{Features from an original image (top), and its shifted variant (bottom). The baseline (ResNet-101) -- without any low-pass filtering, and MaxBlurPool (MBP) \cite{zhang2019making} -- with monotonic blurring, both experience signal degeneration due to aliasing. Our method retains bulk of the expected signal. Here, $28 \times 28$ features are taken from  the res\textsubscript{3} block.}
\label{fig:feats}
\end{figure}

\begin{figure*}
\begin{center}
   \includegraphics[width=.91\linewidth]{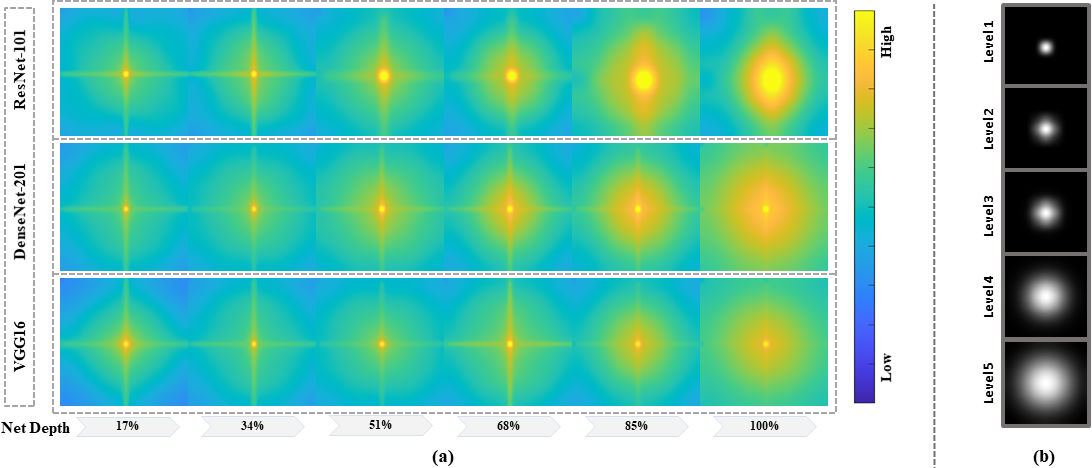}
\end{center}
   \caption{\textbf{(a)} Fourier energy at different network depths (lower frequencies are in the centre). Heat maps are generated using feature maps, at different network depths, belonging to the ImageNet validation set. For all three networks, deeper feature maps have greater high frequency energy, suggesting the importance of depth adaptive blurring. \textbf{(b)} Our proposed method ensures better anti-aliasing by learning stronger blur filters for deeper layers, in this instance, using a ResNet-101 backbone.}
\label{fig:mainCombo}
\end{figure*}

In CNNs, aliasing occurs when down-sampling (DS) operations do not satisfy the Nyquist sampling rate \cite{nyquist1,Azulay1}. The theorem states that if the DS factor is not at least double of the maximum signal frequency, the output will be corrupted. In the context of CNNs, the problem aggravates with increasing network depth, leading to loss of the feature structure, and in effect, loss of shift-invariance. Blurring before sub-sampling -- a well-known signal processing fix, could be effective here \cite{ zhang2019making}, but naively using the same blur kernel across the network, as used in MaxBlurPool (MBP \cite{ zhang2019making}), or using separate ones for each spatial location \cite{zou2020delving} does not yield satisfactory performance. In Figure \ref{fig:feats}, we compare feature maps from three different networks, and find that the baseline (ResNet-101), as well as MBP suffer from signal degeneration.

In this work, we propose a Depth Adaptive Blur-pool (DAB-pool) module and recommend replacing common DS methods with it (details in Section \ref{SubSec:DAB}). DAB is inspired by the Fourier analysis, where we find that the feature maps -- belonging to deeper CNN layers become increasingly higher frequency in nature, as Figure \ref{fig:mainCombo}(a) shows. This heuristic motivates us to let the network learn progressively stronger blur kernels, and Figure \ref{fig:mainCombo}(b) shows such a set of learnt kernels.

Recently, Azulay et al. \cite{Azulay1} have pointed out that aliasing persists even after blurring, as Activation Functions (AFs) often allow high frequency noise to alias back into the signal, and yet, they remain largely unaddressed. As a solution, in this work, we propose a novel AF that not only serves as an effective non-linearity unit, but also acts as a secondary low-pass filter. Similar to DAB-pool, this can also be readily used in existing CNNs for further performance gain.

To assess shift-invariance, we follow benchmark evaluation protocols in \cite{Azulay1,kayhan1,zhang2019making}. We also test our defence against recent (translation-based) adversarial attacks \cite{Engstrom1}, covering both black-box and white-box settings. Although primarily pursuing shift-invariance, we assess performance generalisation on a variety of corruptions and spatial perturbations on ImageNet-C and ImageNet-P respectively \cite{HendrycksALP1}.
In summary, we make the following contributions:

\begin{itemize}
    \item We investigate ways of instilling anti-aliasing properties in CNNs. Since CNNs host a variety of layers, simply inserting a low-pass filter does not suffice. To this end, we propose to redefine the conventional DS and AFs as potential anti-aliasing units in the form of DAB-pool and AA-ReLU respectively.

    \item We analyse the Fourier or spectral properties of deep features and propose to use DAB-pool for DS as the primary anti-aliasing unit.
    
    \item We propose a novel AF with a built-in low-pass filter as a secondary anti-aliasing unit.
    
    \item We use the proposed DS method and AF in VGG16, ResNet-101, and DenseNet-121 by replacing ReLU to demonstrate their effectiveness across architectures.
    
    \item In addition to shift and shift-based perturbations, we also evaluate on adversarial attacks, data corruptions, and image transformations. Our method consistently outperforms contemporary methods across datasets and challenges.
    
\end{itemize}

\begin{figure*}
\begin{center}
   \includegraphics[width=.93\linewidth]{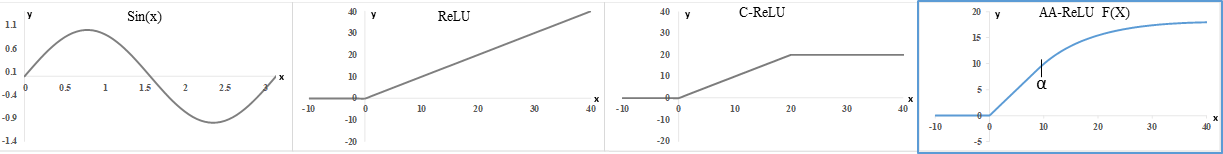}
\end{center}
   \caption{Our proposed AF (AA-ReLU) gates between ReLU and C-ReLU by exploiting the roll-off phase in $\sin x$.}
\label{fig:AFs}
\end{figure*}

\section{Related Work}\label{sec:relatedWork}
\noindent
\textbf{Robust Learning. }
Recently, deep CNN's robustness is put to test in various ways, ranging from adversarial attacks \cite{goodfellow1,madry2018,anguyen1,papernot,moosavi1}, to expose CNNs to common corruptions \cite{akaram1,HendrycksALP1,HendrykRotation01,hossain1,fourier1,geirhos1,RN279,Schneider1}. Latest works \cite{Azulay1,mairal1,fawzi2015,kanbak2018} have shown that even simple image transformations, e.g., shift, can cause misclassification in CNNs. Although data augmentation is effective against such vulnerabilities \cite{augmix,kauderer2017,autoaugment1,cutmix1,cutout1,patchG1,Laptev1}, we delve deeper into the functional blocks of CNN. To be specific, we focus on replacing commonly used DS methods and AFs, largely responsible for shift variance in CNNs \cite{Azulay1}.\\
\\
\textbf{Down-sampling.}
Spatial dimension reduction has been an integral part of CNNs as it reduces the computational overhead and provides local translation invariance \cite{goodfellowBook}. Of late, max-pooling and strided convolution (S-conv) have been predominantly used, owing to their superior task performance \cite{Scherer2}. Nevertheless, these operations do not use a blur-prefix, and in effect, violates the sampling theorem \cite{Azulay1,zhang2019making}. Interestingly, average pooling -- a well known down-sampling method \cite{lecun1}, is effectively a moving blur-filter, and resists aliasing, but cannot match max-pool's superiority in vision tasks \cite{Scherer2}. Upon realising the potential, Zhang et al. \cite{zhang2019making} extended the notion of average pooling, and encouraged the use of stronger filters, e.g., a bilinear or Laplacian \cite{burt1983laplacian}. Here, authors also provided ways to improve the task performance, and yet, used a single filter (MBP) for all layers, which is sub-optimal. 
In SABP \cite{zou2020delving}, authors stressed the importance of an adaptive blur scheme but did not explore the relation with network depth; instead, they used a separate kernel for each local neighbourhood. While this method does alleviate performance, using too many filters per feature map is computationally exhaustive, and more importantly, leads to overfitting. Hypothetically, using a dataset without any location bias could avoid such overfitting; however, in practice, a bias-less dataset hardly exists, e.g., $90\%$ of dog images in ImageNet have the main subject in the centre \cite{Azulay1}, due to a well-known phenomenon -- photographer's bias \cite{Azulay1}. Such bias is prevalent in almost all large-scale datasets, as such, learning separate and spatially local filters do not generalise well (details in Section \ref{sec:results}). 

Kayhan et al. \cite{kayhan1} argued that CNN filters exploit absolute spatial location due to image boundary effect -- a form of convolution irregularity at image borders, and leads to shift variance. Here, an extension of the standard padding scheme is proposed (termed Full-Convolution or F-Conv) to lessen the boundary effects. Nevertheless, performance gain remains marginal.

As an alternative to blurring in the spatial domain, WaveCNet \cite{waveCnets} does so in the frequency domain via Discrete Wavelet Transform (DWT). Here, authors replaced common DS layers with the Low-Low (LL) DWT output. Although effective to some extent, a single DWT low-pass filter is used throughout the network, limiting the overall improvement. Moreover, the back-and-forth wavelet transforms add significant computational overhead. Similar to WaveCNet, Ryu et al. \cite{DFT} also operated in the Fourier domain but only replaced the last average pooling layer with cropped Fourier coefficients (in ResNet). We argue that shift invariance is progressively lost, and only fixing the deepest layer is insufficient. To address this, in an upgraded variant ($DFT^{+}$) \cite{DFT}, Fourier features from shallower layers are extracted in a separate sub-network and fused with the backbone features. Later, SVM is used for final prediction. Even so, $DFT^{+}$ marginally improves performance while incurring a lot of additional parameters. In \cite{Benz}, authors resort to rectifying Batch Normalisation Statistics (BNS) to improve robustness.\\

\noindent
\textbf{Activation Functions. }
AFs inject non-linearity in otherwise linear CNNs, and enable a deep classifier to draw highly non-linear decision boundaries, in high dimensional input space. Rectified Linear Unit (ReLU) is the de-facto AF in modern CNNs \cite{vgg16,resnet1,densenet121}, and although studied for years \cite{relu-father,relu1,relu1Hinton}, its true significance was realized in the last decade, when Krizhevsky et al. \cite{alexkri1,alexnet1} successfully replaced Sigmoidal functions with it. Unlike its predecessors, ReLU offers early sparsity -- a much-desired property that enables training very large and deep networks. Such sparsity gives a network enough `room' to draw accurate decision boundaries among distinct classes. On the flip side though, it lets high frequency noise slip through -- leaving the network vulnerable to aliasing, even with blur routines in place \cite{Azulay1}. Clipped-ReLU (C-ReLU \cite{alexkri1,clipped1}) -- a bounded variant of ReLU, could be used instead, as it caps the output to a constant threshold $\alpha$. However, C-ReLU leaves a sudden and large saturated region beyond $\alpha$, where a derivative is absent. Furthermore, finding an optimal $\alpha$ is difficult as the buffer between the signal and noise is unknown. This highlights the importance of a `soft' capping procedure -- much like how signal rolls-off in a physical low-pass filter \cite{turkowski1990filters}. Different to rectified units, Sigmoidal functions, e.g., hyperbolic \textit{tangent} ($tanh$ \cite{jarrett2009best}), squashes the input and keeps the output within a finite range. 
In \textit{tanh}, the roll-off effect can be regulated as well, but ensuring early feature sparsity remains a problem \cite{alexkri1}. To strike the right balance between early sparsity and noise filtering, in this work, we softly gate between ReLU and C-ReLU using a learnable $\alpha$, as Figure \ref{fig:AFs} shows. 

\section{Proposed Method}\label{sec:proposedMethod}

In this section, we first define the problem set-up. Then we describe DAB-pool and how to integrate it into modern CNNs, followed by the details of our anti-aliasing AF. \\
\\
\textbf{Problem Formulation.}
{Let a CNN classifier} be $\mathcal{G}$ -- a function that takes $X$ as the input and maps it to an output class $\mathcal{O}_{c}$. Considering $X$ has a spatial resolution of $H \times W \times C$, $\mathcal{G}$ can be expressed as $\mathcal{G}(X^{H \times W \times C}) \rightarrow \mathcal{O}_{c}$. For translation invariance to hold, $\mathcal{G}$ should predict the same class label for $X$, and its translated variant as shown in Equation \ref{eq:preli}.

\begin{equation}\label{eq:preli}
\begin{aligned}
\mathcal{G}(X^{H \times W \times C}) & = \mathcal{G}(X^{H \times W \times C}+\Delta {T}) = \mathcal{O}_{c}
\end{aligned}
\end{equation}
where, $\Delta {T}$ represents a marginal input translation.

\subsection{Depth Adaptive Blur-pool (DAB-pool)}\label{SubSec:DAB}

For effective anti-aliasing in CNNs, sufficient high frequencies should be removed before sub-sampling. A signal, containing mostly high frequencies, needs stronger blurring, as opposed to a low frequency signal. This is because a weaker than needed filter might leave frequencies beyond the limit imposed by the Nyquist theorem. On the other hand, using over-aggressive blurring might remove part of the signal itself.

Our spectral analysis, according to Figure \ref{fig:mainCombo}(a), reveals that the feature maps in deeper layers have greater energy quotient for high frequencies. A similar trend is observed in all three networks, which reinforces the consistency of this observation. It also suggests that a `one size fits all' approach, i.e., using the same blur kernel for all layers would be sub-optimal, and hence we propose DAB-pool.

To put these in formal notation, let the feature map for an input $X$ at layer $L$ be $x^{L}$, a Gaussian blur kernel at depth level $D$\footnote{Depth level denotes the number of DS layers, e.g., a network with four DS layers has four depth levels, i.e., $D \in \{1,2,3,4\}$.} be $G^{\sigma}_{D}$, and the output of the convolutional DAB be $Y_{DAB}$. Here, $\sigma$ is the standard deviation of the kernel, and a higher $\sigma$ denotes stronger blurring. Also let the $(L+1)^{th}$ layer be a DS layer and hence we aim to blur $x^{L}$. Now we can define our convolutional DAB operation as Equation \ref{eq:dab} as follows:

\begin{equation}\label{eq:dab}
Y_{DAB}(i, j)=\sum_{p=1}^{m-1} \sum_{q=1}^{n-1} x^{L}_{i+p, j+q} \cdot (G^{\sigma}_{D})_{p, q}
\end{equation}
\noindent
where $Y_{DAB}(i, j)$ is the output at location $(i, j)$ and the Gaussian kernel resolution is $m \times n$.

Rather than using pre-fixed filters \cite{zhang2019making}, we let the network learn them by casting $\sigma$ as a learnable parameter. To ensure progressively stronger blurring with increasing depth $D$, we enforce the following learning constraints:
$G^{\sigma}_{1} < G^{\sigma}_{2} < G^{\sigma}_{3} < G^{\sigma}_{4} \ldots < G^{\sigma}_{M}$,
where $M = max(D)$. \\
\\
\textbf{Replacing Down-sampling Layers.} We replace max-pool, S-conv, and avg-pool as follows:

\begin{equation}\label{eq:maxpool}
\begin{aligned}
\text {Max-pool}_{\textit{k,s}} & \rightarrow \text{Subsample\textsubscript{\textit{k,s}} $\circ$ DAB\textsubscript{\textit{k,s}} $\circ$ DenseMax\textsubscript{\textit{k,s}=1}} \\
& = \text{ DAB-pool\textsubscript{\textit{k,s}} $\circ$ DenseMax\textsubscript{\textit{k,s}=1}}
\end{aligned}
\end{equation}

\begin{equation}\label{eq:conv}
\begin{aligned}
\text {Conv}_{\textit{k,s}>1} & \rightarrow \text{Subsample\textsubscript {\textit{k,s}}$\circ$ DAB\textsubscript {\textit{k,s}} $\circ$ Conv\textsubscript {\textit{k,s}=1}}\\
& = \text{DAB-pool\textsubscript{\textit{k,s}} $\circ$ Conv\textsubscript {\textit{k,s}=1}}
\end{aligned}
\end{equation}

\begin{equation}\label{eq:avg}
\begin{aligned}
\text {AvgPool}_{\textit{k,s}} & \rightarrow \text{Subsample\textsubscript{\textit{k,s}} $\circ$ DAB\textsubscript {\textit{k,s}}} \\
& = \text{DAB-pool\textsubscript{\textit{k,s}}}
\end{aligned}
\end{equation}

\noindent
Here, $k$ and $s$ refer to kernel dimensions and stride respectively. DenseMax is simply stride one max-pooling, found effective in improving task performance \cite{ zhang2019making}. Each sub-sampling operation, in Equations \ref{eq:maxpool}, \ref{eq:conv}, and \ref{eq:avg}, can be easily combined with DAB as a single operation for efficiency, and hence, jointly represented as DAB-pool in the equations.\\
\\
\textbf{Gradient Backpropagation. } In a CNN, partial derivatives of the final cost function are propagated backwards and the learnable parameters are updated accordingly. Here, we demonstrate backpropagation through our proposed DAB layer.

As shown in Equation \ref{eq:dab}, $x^L$ is the input, $G^{\sigma}_{D}$ is the Gaussian blur kernel, and $Y_{DAB}$ is the output. Assume $dy$ is the error that has already backpropagated up to $Y_{DAB}$. From Equation \ref{eq:dab}, we can further derive the following as Equation \ref{eq:backp}.

\begin{equation}\label{eq:backp}
\begin{aligned}
\text{ }Y_{DAB}(i, j) = & \sum_{p=1}^{m-1} \sum_{q=1}^{n-1} x^{L}_{i+p, j+q} \cdot  (G^{\sigma}_{D})_{p, q}\\
\text{Where, }(G^{\sigma}_{D})_{p, q} = &\text{ }\frac{H_{g}\left(p, q\right)}{\sum_{p} \sum_{q} H_{g}}\\
\text{and }H_{g}\left(p, q\right) = &\text{ }e^{\frac{-\left(p^{2}+q^{2}\right)}{2 \sigma^{2}}}
\end{aligned}
\end{equation}

As shown later in Section IV, we use a fixed $3 \times 3$ blur kernel for all DAB layers and hence, $p$ and $q$ always have the same size, i.e., $m = 3$ and $n = 3$. Standard deviation $\sigma$ is the only learnable parameter here that gets updated during training.

During backpropagation, partial derivatives can be calculated by applying the chain rule \cite{chain1}. However, replicating the chain rule via convolution is a more efficient way \cite{backprop1}. Partial derivatives of the weight matrix $G^{\sigma}_{D}$ is calculated by convolving the input $x^L$ with the error matrix $dy$ \cite{backprop1} and represented as $dG^{\sigma}_{D}$. Equation \ref{eq:dw} shows the calculation of $dG^{\sigma}_{D}$. Similarly, error $dy$ is dispersed backwards by convolving it with $G^{\sigma}_{D}$ \cite{backprop1} as shown in Equation \ref{eq:dx}.

\begin{equation}\label{eq:dw}
\begin{aligned}
dG^{\sigma}_{D}(p,q) = & \sum_{p=1}^{m-1} \sum_{q=1}^{n-1}  x^L_{i+p, j+q} \cdot (dy)_{i,j}\\
\end{aligned}
\end{equation}

\begin{equation}\label{eq:dx}
\begin{aligned}
dx^L(i+p,j+q) = & \sum_{p=1}^{m-1} \sum_{q=1}^{n-1} (dy)_{i, j} \cdot (G^{\sigma}_{D})_{p, q}
\end{aligned}
\end{equation}

\subsection{Anti-Aliasing ReLU (AA-ReLU)}\label{SubSec:AAReLU}

In common signal processing tasks, low pass filtering is theoretically sufficient to avoid aliasing. For example, an analogue to digital sound converter inside a voice recorder can guarantee alias-free output, as long as an optimal low-pass filter is used. In contrast, blurring before sub-sampling cannot provide such guarantee in CNNs due to non-linearities in the form of AFs \cite{Azulay1}. We argue that an AF can largely address this by acting as a secondary AA unit that complements DAB-Pool by resisting noise. Such an AF should: allow the signal to go through, have a low-pass filtering or function roll-off mechanism, and function continuity described as below:


\begin{enumerate}[I.]
    \item \textbf{Early feature sparsity \cite{alexkri1}:} Small positive features are mostly true signal \cite{lprelu1} and hence should be allowed to propagate forward -- similar to what C-ReLU offers prior to the clipping point.
    
    \item \textbf{Smooth roll-off:} This is required to complement DAB-pool by high frequency noise suppression. Contrary to a hard-clipping used in C-ReLU, a smooth roll-off is a better choice as the buffer between the signal and noise is unknown. A hard early clipping might affect the original signal, as well as deny early feature sparsity. 
        
    \item \textbf{Function continuity:} For any AF to be effective, it has to be continuous so that there is no point in the function without a derivative.
    
\end{enumerate}
Considering these, we propose an AF in Equation \ref{eq:main} and subsequently describe how it satisfies all three properties.

\begin{equation}\label{eq:main}
F(x)=\left\{\begin{array}{ll}
0, & x \in (-\infty,0] \\

x, & x \in (0,\alpha)\\

f_1^x=\alpha \sin{( \ln \frac{x}{\alpha})}+\alpha, &x \in\left[\alpha, \alpha \exp \left(\frac{\pi}{2}\right)\right]\\

f_2^x=max(f_1^x)= 2\alpha, &x \in\left(\alpha \exp \left(\frac{\pi}{2}\right),+\infty\right)
\end{array}\right.
\end{equation}
where $\alpha$ is a learnable parameter representing the roll-off start point, as well as the amplitude, frequency, and phase shift parameter of our sinusoidal roll-off function $f_1^x$. 

To ensure \textbf{early feature sparsity} \textbf{(\textrm{I})}, the proposed AA-ReLU, as  in Equation \ref{eq:main}, is identical to ReLU for $x \in [- \infty,\alpha)$, however, for $x \geq \alpha$, the signal starts to \textbf{smoothly roll-off} \textbf{(\textrm{II})}. $f_1^x$ dictates the roll-off severity and span, whereas $f_2^x = 2\alpha$ denotes the capping value as shown in Equation \ref{eq:main}.
For further analysis, let $f_1^x$ in Equation \ref{eq:main} be represented as Equation \ref{eq:sinx}:

\begin{equation}\label{eq:sinx}
\begin{aligned}
f_1^x & = \alpha \sin \left(\ln \frac{x}{\alpha}\right)+\alpha \\
& = \alpha \sin (g\left(x\right)) +\alpha, & \text{here, } g(x) = \ln \frac{x}{\alpha}
\end{aligned}
\end{equation}

Unlike the linear growth in ReLU, we aim to roll-off after a certain $\alpha$. To ensure that happens, $f_1^x$ starts by marginally suppressing the output for inputs slightly greater than $\alpha$, i.e., $x > \alpha$. Harder suppression takes effect for sufficiently large $x$, i.e., $x \gg \alpha$. Contrary to the oscillating property of a typical $\sin$ function, $F(x)$, for $x \gg \alpha$, should plateau, i.e., it should reach and stay at the function maximum such that $F(x) = max(F(x)) \{\forall{x} | x \gg \alpha\}$. A $\sin{x}$ function in the interval of $(0,\frac{\pi}{2}]$ has these desirable roll-off properties and our proposed AF in Equation \ref{eq:main} exploits it.

It is worth noting that the input to $f_1^x$ can be large real numbers, i.e., $x \in \mathbb{R}_{>0}$. Therefore, plugging $x$ directly in a $\sin{x}$ function would be impractical, as we do not want our sinusoidal output to oscillate periodically. Rather, we want a substantially lower $\sin$ frequency $f$, so that we have access to a wider and smoother roll-off region. This is why we use a natural logarithmic function $g(x)$ in Equation \ref{eq:sinx}, and in effect, inside our proposed AF in Equation \ref{eq:main}. 

In addition to denoting the roll-off start point, $\alpha$ also increases the amplitude of $f_1^x$ (since $\alpha \in \mathbb{Z}_{>0}$), which allows early sparsity in $F(x)$, as Figure \ref{fig:AFs} shows. $\alpha$ also  vertically shifts $f_1^x$, and further modulates the frequency (by dividing $x$ inside the $\log$ function) such that $F(x)$ remains continuous at each point. 

To show AA-ReLU is \textbf{continuous} \textbf{(\textrm{III})}, we prove that $F(x)$ in Equation \ref{eq:main} is continuous:
\begin{itemize}
    \item $F(x) = 0$ and $F(x) = x$ are continuous since constant and identity functions are continuous everywhere \cite{mathBook}. At the joining point, i.e., for $x = 0$, both the functions produce $0$ and hence they are jointly continuous as well.
    \item $f_1^x$ is continuous in the given domain as it is a function of $\sin$ \cite{mathBook}. At the joining point with $F(x) = x$, i.e., for $x = \alpha$, both functions produce the same output since $f_1^x=\alpha \sin{( \ln \frac{\alpha}{\alpha})}+\alpha = \alpha$ and $F(x) = x$ is an identity function.
    \item $f_2^x$ is another continuous function and at the joining point with $f_1^x$, i.e., for $x = \alpha \exp \left(\frac{\pi}{2}\right)$, $f_1^x=\alpha \sin{( \ln \frac{\alpha \exp \left(\frac{\pi}{2}\right)}{\alpha})}+\alpha = 2\alpha = f_2^x$, and hence, they are also jointly continuous.
\end{itemize}
A case study is presented in the following section for further understanding. \\ 
\\
\textbf{Case Study. } Let a 1D sample signal be $x = $ [-9,  3,  8.9,  9,  10,  43.3,  81]. Assuming $x$ as the input to Equation \ref{eq:main}, and considering $\alpha = 9$, $F(x)$ is evaluated as follows:

\begin{itemize}

    \item $F(-9)$, $F(3)$, and $F(8.9)$ output $0, 3,$ and $8.9$ respectively. By definition, these outputs are identical to ReLU as $x < \alpha$.
    
    \item $F(9) = f_1^9 = 9 \cdot \sin{(\ln \frac{9}{9})} + 9 = 9 \cdot 0 + 9 = 9$ (note $x == \alpha$). From here on, the roll of function smoothly takes over from ReLU. Notice that using the same $\alpha$ as the amplitude, frequency, and phase shift parameter ensures function continuity.
    
    \item $F(10) = f_1^{10} = 9.95$. $x$ has been slightly suppressed since $x > \alpha$.
    
    \item $F(43.3) = f_1^{43.3} = 9 \cdot \sin{(\ln \frac{43.30}{9})} + 9 = 9 \cdot \sin{(\ln e^{1.57})} + 9 = 9 \cdot \sin{\frac{\pi}{2}} + 9 = 18$. $x$ has been heavily suppressed since $x >> \alpha$. 
    
    By calculating the second order derivative of $f_1^x$, we can calculate the function maximum. For this example, solving Equation \ref{eq:second_order} yields $max(F(x)) = 18$.
\begin{equation}\label{eq:second_order}
\begin{aligned}
f_{1}^{x''} & = -\frac{\alpha\left(\sin \left(\ln \left(\frac{x}{\alpha}\right)\right)+\cos \left(\ln \left(\frac{x}{\alpha}\right)\right)\right)}{x^{2}}
\end{aligned}
\end{equation}
    \item $F(81) = f_2^{81} = max(f_1^x) = max(F(x)) = 18$. As shown in Equation \ref{eq:second_order}, $x = 43.3$ induces F(x) to reach its maximum value, and therefore, according to Equation \ref{eq:main}, $ F(81) = F(x>43.3) = 18$.

\end{itemize}

\noindent
\textbf{Gradient. }
Equation \ref{eq:main_grad} shows the derivatives of $F(x)$.

\begin{equation}\label{eq:main_grad}
\frac{d}{dx}F(x)=\left\{\begin{array}{ll}
0, & x \in (-\infty,0], \\
1, & x \in (0,\alpha), \\
\frac{\alpha}{x} \cos{\left( \ln \frac{x}{\alpha}\right)}, &x \in\left[\alpha, \alpha \exp \left(\frac{\pi}{2}\right)\right]\\

0, &x \in\left(\alpha \exp \left(\frac{\pi}{2}\right),+\infty\right)
\end{array}\right.
\end{equation}
Here, beyond $\alpha$, saturated points exist, but an optimally learned $\alpha$ accounts for this issue.

\begin{table*}[]
\centering

\begin{center}
	\resizebox{.85\textwidth}{!}{%
		\begin{tabular}{cccccccccc}

		    \toprule
&&&& \multicolumn{6}{c}{\textbf{Shift Consistency (higher is better)}}    \\  \cmidrule[\heavyrulewidth](lr){5-10}
{\textbf{Backbone}}&{\textbf{Methods}} &
  \multicolumn{2}{c}{\textbf{Clean Accuracy}} &
  \multicolumn{2}{c}{\textbf{Diagonal Shift}} &
  \multicolumn{2}{c}{\textbf{Rescale Shift}} &
  \multicolumn{2}{c}{\textbf{Double Rescaling}} \\ 
\cmidrule[\heavyrulewidth](lr){3-4}\cmidrule[\heavyrulewidth](lr){5-6}\cmidrule[\heavyrulewidth](lr){7-8}\cmidrule[\heavyrulewidth](lr){9-10}
 &  & {Abs} & \textbf{$\Delta$} & {Abs} & \textbf{$\Delta$} & {Abs} & \textbf{$\Delta$} & {Abs} & \textbf{$\Delta$} \\ \cmidrule[\heavyrulewidth](lr){1-10}

& {\cellcolor{gray!20}}Baseline (VGG16) & {\cellcolor{gray!20}}71.66 & - {\cellcolor{gray!20}} & {\cellcolor{gray!20}}88.52 & - {\cellcolor{gray!20}} & {\cellcolor{gray!20}}83.89 & - {\cellcolor{gray!20}} & {\cellcolor{gray!20}}85.04 & - {\cellcolor{gray!20}} \\
 & MBP(Bin-5) \cite{zhang2019making} (ICML '19) & 72.29 & +0.63 & 90.11 & +1.59 & 83.99 & +0.10 & 86.98 & +1.94 \\
   & {\cellcolor{gray!20}}SABP \cite{zou2020delving} (BMVC '20) & {\cellcolor{gray!20}}72.95 & {\cellcolor{gray!20}}+1.29 & {\cellcolor{gray!20}}90.36 & {\cellcolor{gray!20}}+1.84 & {\cellcolor{gray!20}}84.20 & {\cellcolor{gray!20}}+0.31 & {\cellcolor{gray!20}}87.10 & {\cellcolor{gray!20}}+2.06 \\

  VGG16 & BNS \cite{Benz} (WACV '21) & 71.75 & +0.09 & 90.36 & +1.84 & 84.06 & +0.17 & 85.24 & +0.20 \\
     & {\cellcolor{gray!20}}WaveCNet(ch-3,3) \cite{waveCnets} (CVPR '20) & {\cellcolor{gray!20}}71.96 & {\cellcolor{gray!20}}+0.30 & {\cellcolor{gray!20}}90.42 & {\cellcolor{gray!20}}+1.90 & {\cellcolor{gray!20}}85.35 & {\cellcolor{gray!20}}+1.46 & {\cellcolor{gray!20}}86.05 & {\cellcolor{gray!20}}+1.01 \\

& F-Conv \cite{kayhan1} (CVPR '20) & 70.65 & -1.01 & 90.01 & +1.49 & 81.15 & -2.74 & 84.78 & +0.74 \\
 & {\cellcolor{gray!20}}\textbf{{DABP + AA-ReLU (ours)}} & {\cellcolor{gray!20}}\textbf{73.91} & {\cellcolor{gray!20}}+\textbf{2.25} & {\cellcolor{gray!20}}\textbf{91.85} & {\cellcolor{gray!20}}+\textbf{3.33} & {\cellcolor{gray!20}}\textbf{87.47} & {\cellcolor{gray!20}}+\textbf{3.58} & {\cellcolor{gray!20}}\textbf{90.25} & {\cellcolor{gray!20}}+\textbf{5.21}   \\ \cmidrule[\heavyrulewidth](lr){1-10}

& {\cellcolor{gray!20}}Baseline (ResNet-101) & {\cellcolor{gray!20}}77.37 & - {\cellcolor{gray!20}}  &  {\cellcolor{gray!20}}89.95 & - {\cellcolor{gray!20}} & {\cellcolor{gray!20}}80.20  & - {\cellcolor{gray!20}} & {\cellcolor{gray!20}}83.54 & - {\cellcolor{gray!20}} \\
 & MBP(Bin-5) (\cite{zhang2019making}) & 77.40 & +0.03 & 91.31 & +1.36 & 86.10 & +5.90 & 83.97 & +0.43 \\
  & {\cellcolor{gray!20}}SABP (\cite{zou2020delving}) & {\cellcolor{gray!20}}79.32 & {\cellcolor{gray!20}}+1.95 & {\cellcolor{gray!20}}92.19 & {\cellcolor{gray!20}}+2.24 & {\cellcolor{gray!20}}85.94 & {\cellcolor{gray!20}}+5.74 & {\cellcolor{gray!20}}83.81 & {\cellcolor{gray!20}}+0.27 \\

  ResNet-101 & BNS \cite{Benz} & 77.25 & -0.12 & 90.05 & +0.10 & 84.11 & +3.91 & 82.89 & -0.65 \\
     & {\cellcolor{gray!20}}WaveCNet(ch-3,3) \cite{waveCnets} & {\cellcolor{gray!20}}78.21 & {\cellcolor{gray!20}}+0.84 & {\cellcolor{gray!20}}90.89 & {\cellcolor{gray!20}}+0.94 & {\cellcolor{gray!20}}85.95 & {\cellcolor{gray!20}}+5.75 & {\cellcolor{gray!20}}83.15 & {\cellcolor{gray!20}}-0.39 \\

& F-Conv (\cite{kayhan1}) & 78.31 & +0.94 & 90.05 & +0.10 & 84.66 & +4.46 & 82.23 & -1.31 \\
 & {\cellcolor{gray!20}}\textbf{{DABP + AA-ReLU (ours)}} & {\cellcolor{gray!20}}\textbf{81.45} & {\cellcolor{gray!20}}+\textbf{4.08} & {\cellcolor{gray!20}}\textbf{94.11} & {\cellcolor{gray!20}}+\textbf{4.16} & {\cellcolor{gray!20}}\textbf{91.36} & {\cellcolor{gray!20}}+\textbf{11.16} & {\cellcolor{gray!20}}\textbf{86.45} & {\cellcolor{gray!20}}+\textbf{2.91} \\ \cmidrule[\heavyrulewidth](lr){1-10}

 & {\cellcolor{gray!20}}Baseline (DenseNet-121) & {\cellcolor{gray!20}}74.44 & - {\cellcolor{gray!20}} & {\cellcolor{gray!20}}88.81 & - {\cellcolor{gray!20}} & {\cellcolor{gray!20}}82.35 & - {\cellcolor{gray!20}} & {\cellcolor{gray!20}}84.25 & - {\cellcolor{gray!20}} \\
 & MBP(Bin-5) (\cite{zhang2019making}) & 75.03 & +0.59 & 90.39 & +1.58 & 84.21 & +1.86 & 85.74 & +1.49 \\
  & {\cellcolor{gray!20}}SABP (\cite{zou2020delving}) & {\cellcolor{gray!20}}75.36 & {\cellcolor{gray!20}}+0.92 & {\cellcolor{gray!20}}88.85 & {\cellcolor{gray!20}}+0.04 & {\cellcolor{gray!20}}83.55 & {\cellcolor{gray!20}}+1.20 & {\cellcolor{gray!20}}85.77 & {\cellcolor{gray!20}}+1.52 \\

   DenseNet-121 & BNS \cite{Benz} & 74.65 & +0.21 & 88.96 & +0.15 & 82.46 & +0.11 & 85.09 & +0.16 \\
     & {\cellcolor{gray!20}}WaveCNet(ch-3,3) \cite{waveCnets} & {\cellcolor{gray!20}}74.75 & {\cellcolor{gray!20}}+0.31 & {\cellcolor{gray!20}}89.02 & {\cellcolor{gray!20}}+0.21 & {\cellcolor{gray!20}}83.50 & {\cellcolor{gray!20}}+1.15 & {\cellcolor{gray!20}}85.16 & {\cellcolor{gray!20}}+0.91 \\

& F-Conv (\cite{kayhan1}) & 74.09 & -0.35 & 88.84 & +0.03 & 80.88 & -1.47 & 84.55 & +0.03 \\
 & {\cellcolor{gray!20}}\textbf{{DABP + AA-ReLU (ours)}} & {\cellcolor{gray!20}}\textbf{77.44} & {\cellcolor{gray!20}}+\textbf{3.00} & {\cellcolor{gray!20}}\textbf{92.85} & {\cellcolor{gray!20}}+\textbf{4.04} & {\cellcolor{gray!20}}\textbf{88.25} & {\cellcolor{gray!20}}+\textbf{5.90} & {\cellcolor{gray!20}}\textbf{88.08} & {\cellcolor{gray!20}}+\textbf{3.83} 

\\
 \bottomrule
\end{tabular}
}
\end{center}
\caption{Top-1 clean accuracy ($\%$) and shift \textit{consistency} ($\%$) on ImageNet. Our method outperforms others under different shifts. Here, $\Delta$ represents the improvements over baseline and the best result is highlighted in bold.}

\label{table:shift}
\end{table*}

\section{Experiments and Results}\label{sec:results}

We first outline the training details and hyper-parameter settings. Later, for evaluation, we test using: (1) three shift-based perturbations, (2) three recent translation-based adversarial attacks \cite{Engstrom1,Engstrom2}, and (3) a range of corruptions and perturbations. In each set-ups, we refer to the following works for comparison: MaxBlurPool (MBP) \cite{zhang2019making}, Spatially Adaptive Blur Pooling (SABP) \cite{zou2020delving}, Wavelet Integrated CNN (WaveCNet) \cite{waveCnets}, Batch Normalisation Statistics (BNS) \cite{Benz}, and Full-Convolution (F-Conv) \cite{kayhan1}.\\
\\
\subsection{Implementation and Training Details}\label{subsec:implement}
We train and test with three baseline networks -- VGG16 \cite{vgg16}, ResNet-101 \cite{resnet1}, and DenseNet-121 \cite{densenet121}, following the original network architectures presented in respective papers. 
We use Stochastic Gradient Descent ($SGD$) with $Momentum = 0.9$, and train for 100 epochs, while using an initial learning rate $L_r = 0.1$. $L_r$ is dropped by a factor of 0.2 after 30, 60, and 90 epochs and a batch-size of 128 is maintained. Zero-centre input normalisation is used for all experiments, with an L2 Regularization factor of 0.0005. 

As for our learnable parameter $\sigma$, i.e., the standard deviation of Gaussian blur filters, we initialise based on their depth and set $\sigma_{D} = \frac{D}{2}$. 
Take for example ResNet-101: it has five down-sampling layers, and hence, has five depth levels, yielding the following initialisations: $\sigma_{1} = 0.5$, $\sigma_{2} = 1$,  $\sigma_{3} = 1.5$,  $\sigma_{4} = 2$,  and $\sigma_{5} = 2.5$. For all the depth levels, we use $3 \times 3$ blur kernels (see Ablation study in Section \ref{sec:ablation}). 

To initialise the other learnable parameter $\alpha$, placed inside AA-ReLU, we start off with 6 based on heuristic \cite{alexkri1}.

\subsection{Shift Invariance on ImageNet}\label{SubSec:Results_on_Shift}

ImageNet has 1,000 classes with 1.2M training and 50k test images. We follow three shift evaluation protocols from \cite{Azulay1,zhang2019making}.
A network $\mathcal{G}$'s \textit{consistency} (evaluation metric) is the measure of stable Top-1 predictions for an input $X$, and a translated $X_{\Delta T}$, formally put in Equation \ref{eq:consistency}.

\begin{equation}\label{eq:consistency}
\textit{consistency} =\mathbb{E}_{X} \llbracket \{\mathcal{G}\left(X\right) = = \mathcal{G}\left(X_{\Delta T}\right)\} \rrbracket 
\end{equation}

\noindent
where $\llbracket \cdot \rrbracket$ denotes Iverson Bracket that outputs 1 if the proposition inside is true and 0 otherwise. A summary of all three shift perturbation protocols is provided below:\\
\textbf{Diagonal Shift. } here, each $X$ from the validation set is diagonally shifted by $N$ pixels, where $N \in [1,64]$, and $N$ is random. \\
\textbf{Rescale Shift. } Here, each $X$ is first down-scaled to a $100 \times 100$ image, and later, embedded in a $224 \times 224$ canvas, filling rest of the blank space with black pixels. Finally, the resultant image is diagonally shifted by a single pixel to obtain the perturbed variant, i.e.,  $X_{\Delta T}$.\\
\textbf{Double Rescaling. } Here, $X$ is first down-scaled and embedded, similar to Rescale Shift. However, rather than shifting the rescaled embedding itself, it is spatially altered by $\pm1$ pixel to obtain $X_{\Delta T}$ (e.g., from an embedding of $100 \times 100$ to $101 \times 101$).\\

\begin{table}[]
\centering

\begin{center}
	\resizebox{.99\columnwidth}{!}{%
		\begin{tabular}{ccccc}
		\multicolumn{5}{c}{\textbf{Top-1 Accuracy (higher is better)}}   \\ \toprule
&  & \multicolumn{3}{c}{\textbf{Adversarial Attack}} \\
\cmidrule[\heavyrulewidth](lr){3-5} 
\textbf{Methods} & \textbf{Clean} & \textbf{First Order} & \textbf{Grid Search} & \textbf{Worst-of-10} \\
\cmidrule[\heavyrulewidth](lr){1-5}
\rowcolor{gray!20} Baseline (ResNet-101)& 77.37 & 74.78 & 61.50 & 73.39 \\
MBP(Bin-5)(\cite{zhang2019making}) & 77.40 & 74.90 & 63.29 & 73.85 \\
\rowcolor{gray!20} SABP(\cite{zou2020delving}) & 79.32 & 74.96 & 63.55 & 73.44 \\

BNS(\cite{Benz}) & 77.89 & 72.70 & 61.65 & 73.45 \\
\rowcolor{gray!20} WaveCNet(ch-3,3)(\cite{waveCnets}) & 78.05 & 75.05 & 62.30 & 73.66 \\

F-Conv(\cite{kayhan1}) & 78.31 & 73.85 & 62.60 & 72.88 \\
\rowcolor{gray!20} \textbf{DABP+AA-ReLU} & \textbf{81.45} & \textbf{76.28} & \textbf{65.05} & \textbf{74.70} \\ 
\bottomrule
\end{tabular}
}
\end{center}
\caption{Top-1 classification accuracy ($\%$) against different adversarial attacks.}
\label{table:adversarial}
\end{table}

\begin{table*}[]
\centering

\begin{center}
	\resizebox{1\textwidth}{!}{%
		\begin{tabular}{ccccccccccccccccccc}
\textbf{} & \multicolumn{18}{c}{\textbf{ImageNet-C (Corruption Error CE, lower is better)}} \\\cmidrule[\heavyrulewidth](lr){2-19}
\textbf{Methods} & \multicolumn{4}{c}{\textbf{Noise}} & \multicolumn{5}{c}{\textbf{Blur}} & \multicolumn{4}{c}{\textbf{Weather}} & \multicolumn{4}{c}{\textbf{Digital}} & \textbf{} \\
 \cmidrule[\heavyrulewidth](lr){2-5} \cmidrule[\heavyrulewidth](lr){6-10} \cmidrule[\heavyrulewidth](lr){11-14} \cmidrule[\heavyrulewidth](lr){15-18}
\textbf{} & \textbf{Gauss} & \textbf{Shot} & \textbf{Impulse} & \textbf{Speckle} & \textbf{Gauss} & \textbf{Defocus} & \textbf{Glass} & \textbf{Motion} & \textbf{Zoom} & \textbf{Snow} & \textbf{Frost} & \textbf{Fog} & \textbf{Bright} & \textbf{Contr} & \textbf{Elastic} & \textbf{Pixel} & \textbf{Jpeg} & \textbf{mCE} \\ \midrule
\rowcolor{gray!20} Baseline & 68.91 & 72.12 & 74.66 & 73.55 & 64.96 & 61.45 & 74.65 & 60.20 & 61.75 & 66.45 & 60.55 & 54.10 & 30.87 & 60.87 & 54.28 & 54.34 & 44.61 & 61.08 \\
MBP(Bin-5) (\cite{zhang2019making}) & 64.45 & 66.05 & 68.98 & 71.56 & 62.34 & 61.06 & 70.30 & 61.25 & 60.35 & 66.01 & 58.21 & 50.24 & 29.91 & 59.87 & 55.36 & 48.75 & 39.54 & 58.48 \\
\rowcolor{gray!20} SABP (\cite{zou2020delving}) & 65.32 & 65.87 & 69.50 & 71.10 & 63.80 & 61.11 & 68.32 & 60.05 & 60.14 & 64.98 & 58.33 & 49.87 & 29.87 & 59.70 & 53.88 & 46.68 & 38.22 & 58.04 \\

BNS (\cite{Benz}) & 67.11 & 69.89 & 70.98 & 74.52 & 67.23 & 64.90 & 73.85 & 67.21 & 62.92 & 71.66 & 63.24 & 54.87 & 33.87 & 60.45 & 57.32 & 52.82 & 43.75 & 62.15 \\
\rowcolor{gray!20} WaveCNet(ch-3,3) (\cite{waveCnets}) & 66.32 & 69.15 & 70.08 & 74.14 & 66.14 & 63.36 & 71.98 & 65.05 & 62.44 & 69.70 & 61.34 & 54.01 & 31.22 & 60.14 & 56.99 & 52.32 & 41.66 & 60.94\\

F-Conv (\cite{kayhan1}) & 66.98 & 65.21 & 69.37 & 72.88 & 62.77 & 60.81 & 72.66 & 60.32 & 61.45 & 66.02 & 60.24 & 52.36 & 29.56 & 59.86 & 54.44 & 46.87 & 39.35 & 58.89 \\
\rowcolor{gray!20} \textbf{DABP+AA-ReLU} & \textbf{61.45} &  \textbf{62.19} &  \textbf{65.32} &  \textbf{68.95} &  \textbf{59.87} &  \textbf{60.77} &  \textbf{66.25} &  \textbf{60.01} &  \textbf{59.98} &  \textbf{64.21} &  \textbf{58.12} &  \textbf{47.87} &  \textbf{29.41} &  \textbf{59.35} &  \textbf{53.21} &  \textbf{44.24} &  \textbf{36.97} &  \textbf{56.36}\\ \bottomrule
\end{tabular}
}
\end{center}
\caption{Corruption Error rate ($\%$) on ImageNet-C \cite{HendrycksALP1} (Baseline: ResNet-101). Results in each category represent the average across all five severity levels in the dataset. The rightmost column (mCE) denotes the mean Corruption Error.}
\label{table:corruption}
\end{table*}

\noindent
\textbf{Results. }
Compared to the baseline (VGG16, ResNet-101 and DenseNet-121), DABP+AA-ReLU improves shift-invariance across all three deep networks in Table \ref{table:shift} despite substantial architectural difference. For instance, VGG16 has four intermediate max-pool layers as opposed to three average pooling layers in DenseNet-121. On the other hand, ResNet-101 and DenseNet-121 use residual block and depth concatenation, respectively, whereas VGG16's architecture is much simpler with sequential input and output layers. Interestingly, our method improves the clean Top-1 accuracy as well exceeding the baseline by $+2.25\%$ in VGG16, by $+4.08\%$ in ResNet-101, and by $+3\%$ in DenseNet-121. Compared to a fixed blur kernel in MBP \cite{zhang2019making} and a spatially adaptive blur kernel in SABP \cite{zou2020delving}, 
DABP+AA-ReLU shows better consistency in each of the evaluation protocols in Table \ref{table:shift}. Unlike other methods, it addresses the role of AFs as well, which complements DAB-pool in achieving strong shift-invariance.

\subsection{Robustness Against Adversarial Attacks}\label{SubSec:results_on_Attack}
By definition, an image $X'$ is an adversarial variant of some valid image $X$, if both appear visually similar to a human, and yet, a network misclassifies $X'$. Although visual perception is subjective and similarity between images is hard to quantify, the relevant literature presumes that $X'$ is adversarial if and only if 
$\left\|X-X^{\prime}\right\|_{p} \leq \varepsilon$, where $p \in [1,\infty)$ and $\varepsilon$ is small. Traditional first-order attacks, e.g., Fast Gradient Sign Method (FGSM \cite{goodfellow1}),  operate in pixel space and generate adversaries from a bounded $\ell_{p}$ ball. As recently pointed out in \cite{Engstrom1}, spatial adversarial vulnerability, e.g., due to translation, do not obey the bounded protocol. This is because a translated image $X_{\Delta T}$ can still appear visually similar, and yet, have a substantially large norm $\left\|X-X^{\prime}\right\|_{p}$, compared to the norms allowed in $\ell_{p}$. Hence, finding adversaries within a confined $\ell_{p}$ space is difficult with first-order methods. In this work, we adhere to the spatial attacks outlined in \cite{Engstrom1} and summarised below:\\
\\
\textbf{First Order Method (FO). }
FO is a white-box attack, meaning it requires access to the target network architecture, and in effect, to its gradients. In $\ell_{p}$ based adversaries, each input pixel is cast as a learnable parameter in a bid to maximise loss for a particular class (to inflict misclassification). However, here, FO has to perturb the spatial tuning of an image $X$ to generate a translated adversary $X'$. This is why input translation is parameterised in FO, and the loss for the correct class is maximised by iteratively updating these parameters. Following \cite{Engstrom1}, the maximum perturbation, i.e., translation, allowed in this attack is 25 pixels, and in terms of implementation, it is 200 steps of projected gradient descent with step size 0.24.\\
\\
\textbf{Grid Search Attack (GSA). }
This is a black-box attack and requires only a few queries to the target model. GSA attacks the network by exhaustively translating $X$ in each spatial direction until $X$ becomes an adversary $X'$ (if such an $X'$ exists in the search grid). Here, maximum translation allowed in each direction is 5 pixels.\\
\\
\textbf{Worst-of-$k$. }
Here, to attack a target, $X$ is translated randomly $k$ times by keeping the translation parameters within the allowed limit, i.e., 25 pixels. An $X'$, within this limit, inducing the worst classification performance is the one chosen as the adversary. For example, in worst-of-10 attack, it performs 10 random translations per $X$ and feeds the corresponding $X'$s to find the most effective adversary. It is a black-box attack as well.\\
\\
\textbf{Results. }
As Table \ref{table:adversarial} shows, our method exhibits better resistance against all three attacks in terms of Top-1 accuracy. Interestingly, DABP+AA-ReLU performs best against FO -- the most effective attack in pixel-level $\ell_{p}$ bounded space, performing $1.5\%$ over baseline (from here on, we refer to ResNet-101 as the baseline unless mentioned otherwise). This can be attributed to different loss landscapes in natural image transformations versus typical $\ell_{p}$ settings \cite{Engstrom1}. In $\ell_{p}$ loss landscape, the maxima are consistent and concentrate well in a locality, making it easier to find adversaries by taking small iterative steps towards it \cite{madry2018,Mohapatra1}. However, the loss landscape of translation is non-convex and tend to have multiple scattered local maxima \cite{Engstrom1}, meaning there is no guarantee that taking small steps would lead to translated adversaries, and hence, it is sub-optimal. GSA, on the other hand, turns out to be more challenging, and the baseline accuracy drops from $77.4\%$ to $61.5\%$. DABP+AA-ReLU performs better than MBP by $1.76\%$, SABP by $1.50\%$, BNS by $3.40\%$, WaveCNet by $2.75\%$, and F-Conv by $2.45\%$ under GSA. A similar improvement trend is observed against Worst-of-10 attack as well.

\begin{figure}
\begin{center}
   \includegraphics[width=.85\linewidth]{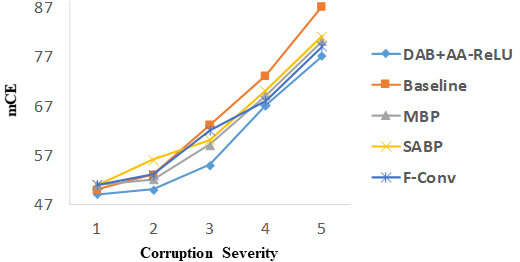}
\end{center}
   \caption{Mean Corruption Error (\%) on increasing corruption severity levels in ImageNet-C (lower is better).}
\label{fig:severity}
\end{figure}

\subsection{Corruption and Perturbation Robustness}\label{SubSec:results_on_INC_INP}

In this section, we test our corruption and perturbation stability on ImageNet-C and ImageNet-P respectively.\\
\textbf{ImageNet-C. }
ImageNet-C \cite{HendrycksALP1} contains images from the ImageNet validation set with 17 different corruptions, broadly falling into the following four categories: noise, blur, weather, and digital. Each of the 17 corruptions has five severity levels and mean Corruption Error (mCE) is used as the evaluation metric.\\
\\
\textbf{Results. }
Table \ref{table:corruption} shows that DABP+AA-ReLU achieves state-of-the-art robustness against a variety of corruptions in ImageNet-C with an mCE $4.71\%$ lower than the Baseline, $2.12\%$ lower than MBP \cite{zhang2019making} with Binomial-5 filter, $1.68\%$ than SABP \cite{zou2020delving}, $5.79\%$ than BNS \cite{Benz}, $4.58\%$ than WaveCNet with cohen-3,3 filter, and $2.53\%$ than F-Conv. Our method exhibits marked robustness in the noise category consisting of Gaussian, shot, impulse, and speckle noise, achieving an overall $7.83\%$ lower mCE than the baseline. Compared to competing methods, our overall noise mCE is $3.28\%$ lower than MBP \cite{zhang2019making} and $3.47\%$ lower than SABP \cite{zou2020delving}. This is an expected byproduct of blurred DS as most of the high frequency noise gets neutralised in this step. However, signal noise -- much like aliasing, can reappear in the non-linear AFs, which we believe is the case in the baseline and in \cite{zhang2019making,zou2020delving}. Our method is better suited to counter this challenge as AA-ReLU itself acts as a secondary low-pass filter. The baseline often performs well under level-1 corruption severity, but struggles beyond level 2 and 3 (see Figure \ref{fig:severity}). Our method shows better consistency, even with increasing corruptions.\\
\\
\textbf{ImageNet-P. }
ImageNet-P \cite{HendrycksALP1} introduces a range of perturbations in the validation set to test performance stability. In this work, we evaluate on four affine transformation-based perturbations involving translation, rotation, scaling, and tilt. In ImageNet-P, each test instance is a short video of about 30 frames with increasing perturbation severity. Flip Probability FP is used as the stability evaluation metric. A flip event occurs when two consecutive frames' predictions mismatch. To put it formally, let us denote $k$ perturbation sequences (each with $v$ number of frames) with $S = \left\{\left(x_{1}^{(i)}, x_{2}^{(i)}, \ldots, x_{v}^{(i)}\right)\right\}_{i=1}^{k} $.
For a fixed $i$, i.e., perturbation type $k$, $x_{1}^{(i)}$ denotes the clean image (no perturbation) and $x_{v}^{(i)}$ denotes a frame with maximum ($k$ type) perturbation. The Flip Probability (FP) of a deep classifier $\mathcal{G}$ on perturbation sequence $S$ is:
\begin{equation}
F P_{p}^{\mathcal{G}} =\frac{1}{k(l-1)} \sum_{i=1}^{k} \sum_{j=2}^{v} 1\left(\mathcal{G}\left(x_{j}^{(i)}\right) \neq \mathcal{G}\left(x_{j-1}^{(i)}\right)\right) 
\end{equation}
\textbf{Results. }
As Table \ref{table:perturbation} shows, our method shows better stability ($0.87\%$ lower mFP than the next best SABP), and interestingly, it generalises to perturbations other than translation as well, e.g., in rotation perturbation, our FP is $2.27\%$ lower than the baseline, $1.34\%$ than MBP, and $1.17\%$ than SABP; gain is observed for tilt and scale as well.

\begin{table}[]
\centering

\begin{center}
	\resizebox{1\columnwidth}{!}{%
		\begin{tabular}{cccccc}
                    \multicolumn{6}{c}{\textbf{ImageNet-P Flip Probability FP (lower is better)}}   \\ \cmidrule[\heavyrulewidth](lr){1-6}
           & \multicolumn{5}{c}{\textbf{Affine Transform Perturbation}}     \\ \cmidrule[\heavyrulewidth](lr){2-6}
 \textbf{Methods}                   & \textbf{Translate}       & \textbf{Rotate}        & \textbf{Tilt}    & \textbf{Scale}     & \textbf{mFP}          \\ \cmidrule[\heavyrulewidth](lr){1-6}
\rowcolor{gray!20} Baseline           & 4.15                                  & 6.39                                  & 3.96                                  & 10.85                                 & 6.34                                  \\
MBP(Bin-5) (\cite{zhang2019making})        & 3.70                                  & 5.46                                  & 3.37                                  & 8.41                                  & 5.24                                  \\
\rowcolor{gray!20} SABP (\cite{zou2020delving})         & 3.42                                  & 5.29                                  & 3.32                                  & 8.06                                  & 5.02                                  \\

BNS (\cite{Benz})        & 4.46                                  & 5.98                                  & 4.20                                  & 9.35                                  & 6.00                                  \\
\rowcolor{gray!20} WaveCNet (\cite{waveCnets})         & 3.58                                  & 5.87                                  & 3.68                                  & 8.89                                  & 5.51                                  \\

F-Conv (\cite{kayhan1})    & 4.15                                  & 6.74                                  & 4.25                                  & 9.01                                  & 6.04                                 \\
\rowcolor{gray!20} \textbf{DABP+AA-ReLU} &   \textbf{2.26} &   \textbf{4.12} &   \textbf{3.22} &   \textbf{6.98} &   \textbf{4.15} \\
\bottomrule
\end{tabular}
}
\end{center}
\caption{Flip Probability rate ($\%$) on transformation-based perturbations in ImageNet-P \cite{HendrycksALP1} (Baseline: ResNet-101). mFP denotes mean Flip Probability.}
\label{table:perturbation}
\end{table}

\begin{table}[!ht]

	\label{t4}
	\begin{center}
	\resizebox{1\columnwidth}{!}{%
		\begin{tabular}{c|ccc}
 \hline
  Network           & Time/epoch (mins) & No. of epochs\\ \hline
 ReLU+(S-conv, Max-pool) (baseline)              & 16.3    &  $[105,115]$  \\
 C-ReLU+(S-conv, Max-pool)      & 18.2     &  $[85,95]$ \\
 ReLU+SABP \cite{zou2020delving}            & 24.9     &  $[100,110]$\\
 AA-ReLU+DAB-pool (Ours) & 22.6  &  $[90,100]$    \\
\hline
\end{tabular}
}
\end{center}
\caption{Comparison of training time (on ImageNet). Average time required to complete an epoch is presented in minutes and the corresponding number of epochs for convergence is shown within a range. ResNet-101 is used as the baseline in all the networks.}
\label{table:time}
\end{table}

\subsection{Training Time}

By definition, AA-ReLU executes more conditional statements than ReLU during training. Similarly, our proposed DAB-pool layer has an additional blurring operation compared to Strided Convolution (S-conv) or Max-pool with a single operational stage. Therefore,  our CNN requires more training time compared to the baseline, i.e., vanilla ResNet-101 with ReLU + (S-conv and Max-pool).
As shown in Table \ref{table:time}, this baseline is the fastest with 16.3 mins/epoch. ResNet-101 with C-ReLU is slightly slower with 18.2 mins/epoch. Our network takes 22.6 mins/epoch which is 2.3 minutes lesser than the SABP network. It should be noted that our method requires $[90,100]$ training epochs to converge, whereas SABP and the baseline require more epochs to converge. We argue that the unbounded nature of ReLU results in additional training iterations for convergence. In contrast, AA-ReLU has a bounded output and hence, the search space for the global minima is much smaller. This means our network requires fewer iterations of weight updating to reach this minima resulting in a reduced convergence time. Similar to AA-ReLU, C-ReLU also has faster convergence due to its bounded nature. 

All the training hyper-parameters are same as discussed earlier in Section \ref{subsec:implement}.

\section{Ablation Study}\label{sec:ablation}
To justify our design choices, we conduct an ablation study on ImageNet using ResNet-101 backbone. All the other settings remain identical to that of Section \ref{sec:results}.

As Table \ref{table:ablation} shows, a $3 \times 3$ Gaussian blur filter performs better than $5 \times 5$ and $7 \times 7$ filters. We hypothesise that too big of a blur kernel spreads the features across an unnecessarily large region. This is particularly concerning for deeper layers where filters have large receptive fields, e.g., in ResNet-101, a $224 \times 224$ input reduces down to $14 \times 14$ feature maps in the last DS layer. Using a $7 \times 7$ filter which is half as big as the feature map dimension turns out to be sub-optimal. 

We also test the compatibility of different AFs with DAB-pool and max-pool. Although C-ReLU performs better than ReLU under all three shifts, it lags behind in clean accuracy. We argue that because of C-ReLU's bounded nature, it has some anti-aliasing property, but a large saturated region abstains it from achieving higher clean accuracy. Overall, DABP\textsubscript{3*3}+AA-ReLU is the best configuration achieving $2.41\%$ and $2.42 \%$ higher average accuracy than DABP\textsubscript{3*3}+C-ReLU and DAB\textsubscript{3*3}+ReLU  respectively.

\begin{table}[!ht]
\centering

\begin{center}
	\resizebox{.99\columnwidth}{!}{%
		\begin{tabular}{ccccc}
		\multicolumn{5}{c}{\textbf{Ablation Study (ImageNet)}}   \\ 
		\toprule
		&& \multicolumn{3}{c}{\textbf{Shift Consistency (higher is better)}} \\
		\cmidrule[\heavyrulewidth](lr){3-5}
 & \textbf{Clean} & \textbf{Diagonal} & \textbf{Rescale} & \textbf{Double} \\ 
\textbf{Methods} & \textbf{Accuracy} & \textbf{shift} & \textbf{shift} & \textbf{Rescaling} \\
\toprule
\rowcolor{gray!20} DABP\textsubscript{3*3}+ReLU & 79.95 & 91.19 & 88.41 & 84.15 \\
DABP\textsubscript{5*5}+ReLU & 78.85 & 88.54 & 87.37 & 83.15 \\
\rowcolor{gray!20} DABP\textsubscript{7*7}+ReLU & 77.03 & 85.21 & 83.87 & 81.29 \\
DABP\textsubscript{3*3}+C-ReLU & 79.03 & 91.58 & 88.80 & 84.33 \\
\rowcolor{gray!20} DABP\textsubscript{5*5}+C-ReLU & 78.25 & 88.75 & 87.54 & 83.36 \\
DABP\textsubscript{7*7}+C-ReLU & 74.87 & 87.25 & 86.66 & 81.89 \\
\rowcolor{gray!20} \textbf{DABP\textsubscript{3*3}+AA-ReLU} & \textbf{81.45} & \textbf{94.11} & \textbf{91.36} & \textbf{86.45} \\
DABP\textsubscript{5*5}+AA-ReLU & 78.94 & 90.66 & 88.96 & 84.90 \\
\rowcolor{gray!20} DABP\textsubscript{7*7}+AA-ReLU & 76.15 & 88.96 & 87.32 & 83.08 \\
Max-pool+ReLU & 76.37 & 88.95 & 79.20 & 82.54 \\
\rowcolor{gray!20} Max-pool+C-ReLU & 75.01 & 88.98 & 79.89 & 83.04 \\
Max-pool+AA-ReLU & 76.90 & 89.06 & 89.23 & 83.85 \\ 
\bottomrule
\end{tabular}
}
\end{center}
\caption{Ablation study under varying architectural settings.}
\label{table:ablation}
\end{table}

\section{Conclusion}
To avoid aliasing and make CNNs invariant to small input changes, we have shown that benefits can be harnessed if depth is considered for blurred down-sampling. To this end, we have introduced DAB-pool, and outlined ways to replace common down-sampling operations with it. To ensure the problem does not reappear, we have proposed a new activation function AA-ReLU, which also serves as a low-pass filter.  
We have evaluated against a broad array of shift, adversarial attack, corruption, and perturbation, and demonstrated the efficacy of our method under challenging conditions. In doing so, we have also integrated our modules in three different networks and shown that the benefits generalise well despite architectural differences. 
Since classification network backbones are an integral part of many more vision tasks such as object detection and semantic segmentation, we believe our proposed methods are easily transferable to these tasks as well.

\ifCLASSOPTIONcaptionsoff
  \newpage
\fi

{
\bibliographystyle{IEEEtran}
\bibliography{egbib}
}

\vspace{-10 mm}
\begin{IEEEbiography}[{\includegraphics[width=1in,height=1.25in,clip]{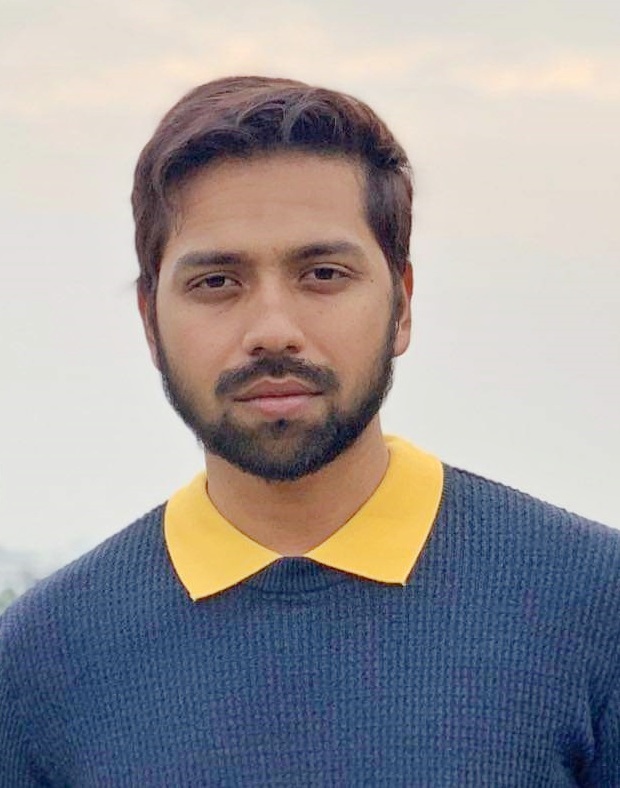}}]{Md Tahmid Hossain} is currently pursuing his Ph.D. degree at the School of Science, Engineering and Information Technology, Federation University, Australia. He received a Bachelor of Science (B.Sc.) degree in Computer Science and Engineering (CSE) from the Islamic University of Technology (IUT), Bangladesh, in 2015. He has held the position of lecturer in the department of CSE at the same university he graduated from. His area of research interests includes deep learning-based image classification, object detection, generative adversarial networks and computer vision in general.
\end{IEEEbiography}
\vspace{-10 mm}
\begin{IEEEbiography}[{\includegraphics[width=1in,height=1.25in,clip]{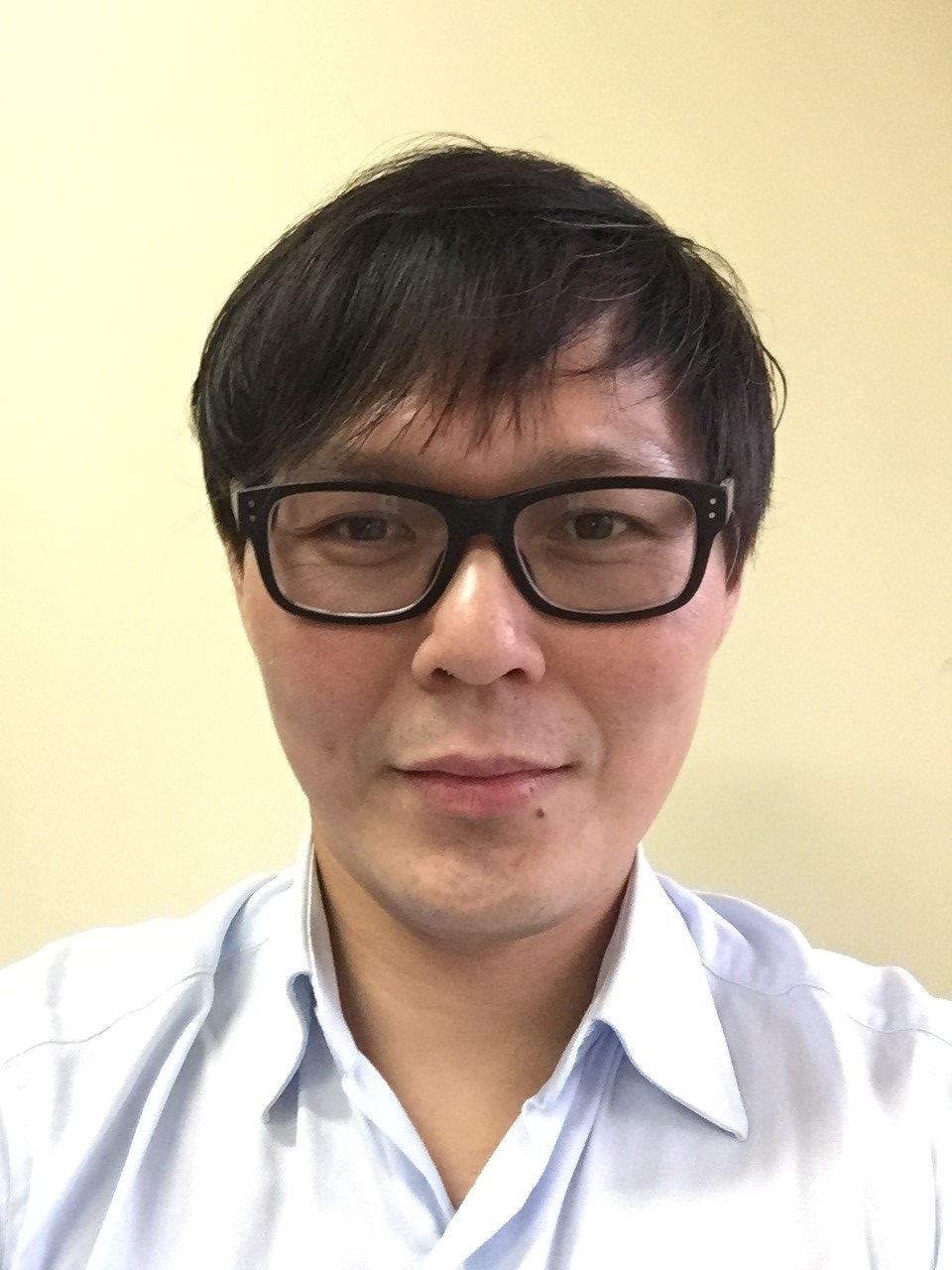}}]{Shyh Wei Teng} is an Associate Professor and Deputy Dean at School of Engineering, Information Technology and Physical Sciences, Federation University Australia. He previously held positions at Monash University after he obtained his PhD at that university in 2004. His research interests include Image/video processing; Machine learning; and Multimedia analytics. He has so far published over 80 refereed research papers. He received various competitive research funding, including three Federal Government grants on image retrieval and analytics—one from the Australian Research Council (ARC). He supervised 9 PhD and 1 MPhil students to completion.
\end{IEEEbiography}
\vspace{-10 mm}

\begin{IEEEbiography}[{\includegraphics[width=1in,height=1.25in,clip]{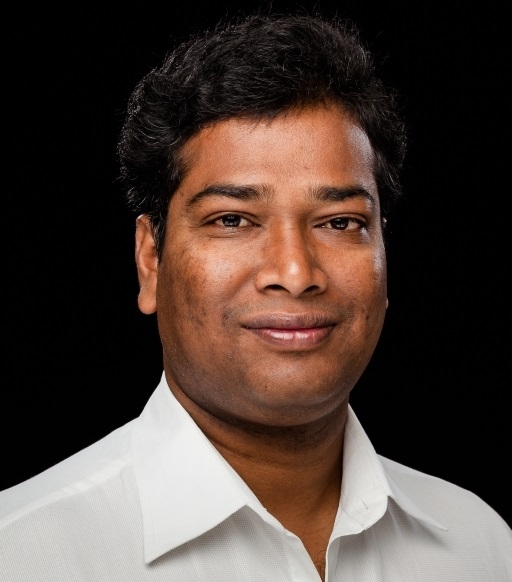}}]{Ferdous Sohel} (M’08-SM’13) received a PhD degree from Monash University, Australia. He is currently an Associate Professor in Information Technology at Murdoch University, Australia. Prior to joining Murdoch, he was a Research Fellow at the School of Computer Science and Software Engineering, The University of Western Australia from January 2008 to 2015. His research interests include computer vision, machine learning, pattern recognition, and digital agriculture. He is a recipient of the best PhD thesis medal from Monash University. He has received several best paper awards. He was a co-presenter of a tutorial at CVPR2015. He is an Associate Editor of IEEE Transactions on Multimedia (2020-2021) and IEEE Signal Processing Letters (2020-2021). He served as a Technical Program Chair of DICTA2021, DICTA2019; as an Organising Secretary of APCC2017; and as a Tutorial Chair of PSIVT2019. He is a member of the Australian Computer Society and a senior member of IEEE.

\end{IEEEbiography}
\vspace{-12 mm}

\begin{IEEEbiography}[{\includegraphics[width=1in,height=1.25in,clip]{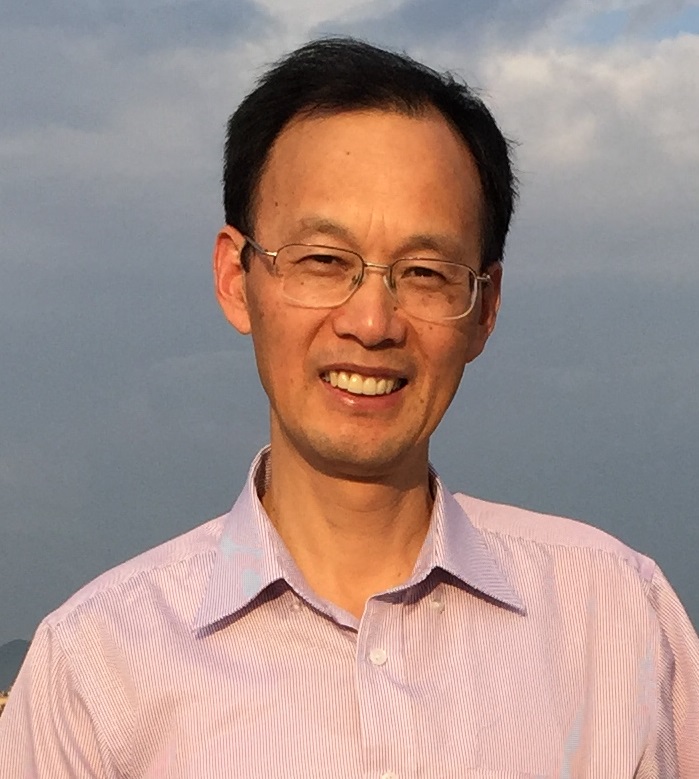}}]{ Guojun Lu} (Senior Member of IEEE) is a Professor at School of Engineering, Information Technology and Physical Sciences, Federation University Australia. He has many years’ research experience in artificial intelligence, multimedia signal processing, and retrieval, and has worked on an ARC DP project and supervising an ARC DECRA project. He has successfully supervised over 20 PhD students. He has held positions at Loughborough University, National University of Singapore, Deakin University and Monash University, after he obtained his PhD in 1990 from Loughborough University and BEng from Nanjing Institute of Technology (now South East University, China). He has published over 230 refereed journal and conference papers and wrote two books( Communication and Computing for Distributed Multimedia Systems (Artech House 1996), and Multimedia Database Management Systems (Artech House 1999)).
\end{IEEEbiography}

\end{document}